\documentclass{ieeeaccess}
\usepackage{cite}
\usepackage{amsmath,amssymb,amsfonts}
\usepackage{graphicx}
\usepackage{textcomp}

\usepackage{algorithm}
\usepackage{algpseudocode}
\algnewcommand{\Input}[1]{\item[\textbf{Input:}] #1}
\algnewcommand{\Output}[1]{\item[\textbf{Output:}] #1}

\usepackage{booktabs,tabularx,longtable,array,multirow,arydshln}
\usepackage{framed}
\usepackage{listings}
\usepackage{bm}
\makeatletter
\AtBeginDocument{\DeclareMathVersion{bold}
\SetSymbolFont{operators}{bold}{T1}{times}{b}{n}
\SetSymbolFont{NewLetters}{bold}{T1}{times}{b}{it}
\SetMathAlphabet{\mathrm}{bold}{T1}{times}{b}{n}
\SetMathAlphabet{\mathit}{bold}{T1}{times}{b}{it}
\SetMathAlphabet{\mathbf}{bold}{T1}{times}{b}{n}
\SetMathAlphabet{\mathtt}{bold}{OT1}{pcr}{b}{n}
\SetSymbolFont{symbols}{bold}{OMS}{cmsy}{b}{n}
\renewcommand\boldmath{\@nomath\boldmath\mathversion{bold}}}
\makeatother

\def\BibTeX{{\rm B\kern-.05em{\sc i\kern-.025em b}\kern-.08em
    T\kern-.1667em\lower.7ex\hbox{E}\kern-.125emX}}

\begin{document}
\doi{}

\title{MindFlow+: A Self-Evolving Agent \\for E-Commerce Customer Service}
\author{{Ming Gong}\authorrefmark{1,2}, {Xucheng Huang}\authorrefmark{1}, {Ziheng Xu}\authorrefmark{1}, {Vijayan K. Asari}\authorrefmark{2}}

\address[1]{Xiaoduo AI Lab, Shanghai, China}
\address[2]{University of Dayton, Dayton, Ohio, United States}

\markboth
{Ming Gong \headeretal: MindFlow+: A Self-Evolving Agent for E-Commerce Customer Service}
{Ming Gong \headeretal: MindFlow+: A Self-Evolving Agent for E-Commerce Customer Service}

\corresp{Corresponding author: Ming Gong (e-mail: gongm1@udayton.edu).}

\begin{abstract}
High-quality dialogue is crucial for e-commerce customer service, yet traditional intent-based systems struggle with dynamic, multi-turn interactions. We present MindFlow+, a self-evolving dialogue agent that learns domain-specific behavior by combining large language models (LLMs) with imitation learning and offline reinforcement learning (RL). MindFlow+ introduces two data-centric mechanisms to guide learning: tool-augmented demonstration construction, which exposes the model to knowledge-enhanced and agentic (ReAct-style) interactions for effective tool use; and reward-conditioned data modeling, which aligns responses with task-specific goals using reward signals. To evaluate the model’s role in response generation, we introduce the AI Contribution Ratio, a novel metric quantifying AI involvement in dialogue. Experiments on real-world e-commerce conversations show that MindFlow+ outperforms strong baselines in contextual relevance, flexibility, and task accuracy. These results demonstrate the potential of combining LLMs, tool reasoning, and reward-guided learning to build domain-specialized, context-aware dialogue systems.
\end{abstract}


\titlepgskip=-21pt
\maketitle

\section{Introduction}

The rapid growth of e-commerce has made it an essential part of modern life. As the industry evolves, customer service emerges as a critical competitive factor, directly influencing user satisfaction and business outcomes~\cite{gajewska2020impact}. To scale support efficiently, many businesses are turning to automation~\cite{chaturvedi2023opportunities, ren2024survey}. However, traditional NLP-based systems, centered on intent recognition and template-based responses, struggle to handle the diverse, dynamic, and often ambiguous nature of real-world e-commerce queries.

In parallel, Large Language Models (LLMs) have demonstrated impressive capabilities across a wide range of tasks~\cite{zhao2024surveylargelanguagemodels}. However, their direct application in e-commerce remains limited. Unlike static knowledge retrieval or open-domain chat, e-commerce scenarios demand domain-specific understanding, real-time adaptability, factual grounding, and the ability to make multi-step decisions based on evolving user intent. These requirements exceed the capabilities of general-purpose dialogue systems~\cite{bamberger2023generative}.

To bridge this gap, we introduce MindFlow+, a self-evolving agent specifically designed for e-commerce dialogue automation. Unlike traditional intent-based systems, MindFlow+ is capable of adapting to diverse customer needs through dynamic behavior learning.

MindFlow+ learns domain-specific behavior through supervised fine-tuning (SFT) on interaction data enriched with tool calls, external knowledge, and reward-guided supervision. This approach allows the agent to internalize task-specific patterns, reason over external tools, and generate responses that are both contextually grounded and aligned with human preferences. By unifying tool-use reasoning with preference-aligned generation in a single training framework, MindFlow+ achieves adaptive behavior without requiring architectural modifications to the base LLM.

The primary contributions of this research are summarized as follows:

\begin{enumerate}
\item \textbf{MindFlow+}, a self-evolving agent framework for e-commerce dialogue automation that performs SFT on integrated interaction data enriched with tool calls, external knowledge, and reward-guided supervision, unifying tool-augmented reasoning and preference-aligned response generation into a single training process, enabling adaptive and context-aware responses without modifying the underlying large language model architecture.

\item \textbf{AI Contribution Ratio}, a novel metric to quantify the agent’s autonomy in multi-step workflows. It captures both decision-making efficiency and the model’s practical contribution to real-world tasks, offering fine-grained insight into agent capability.

\item \textbf{Expansion to Broader Applications in LLM-based Dialogue Systems}, a versatile approach that extends MindFlow+’s adaptability and contextual understanding beyond e-commerce, enabling improved user interactions in diverse domains such as virtual assistants, help desks, and other task-oriented conversational AI systems.

\end{enumerate}

\section{Related Work}

\subsection{NLP in E-Commerce Customer Service}
In the early stages of e-commerce customer service, question-answering (QA) systems were based on rule-based methods, like predefined FAQs and keyword matching~\cite{weizenbaum1966eliza, malik2024natural}. Although these methods offered basic automation, they struggled with complex issues and multi-turn dialogues, limiting their adaptability.

With advancements in deep learning and NLP, rule-based systems have been largely replaced by retrieval- and generation-based models, improving dialogue quality and accuracy~\cite{mashaabi2022natural, olujimi2023nlp, khurana2023natural}. Modern systems use diverse data sources such as product reviews and past QA pairs to improve answer generation~\cite{gao2021meaningful, yu2018responding}, with knowledge graphs further enhancing retrieval accuracy~\cite{xu2024retrieval, tapeh2008knowledge}. Scalable solutions that utilize user-generated content also help address repetitive queries~\cite{cui2017superagent, mittal2021distantly}. Despite these advancements, challenges remain in addressing complex, context-dependent queries and managing multi-turn interactions.

\subsection{LLMs in E-Commerce Customer Service}

LLMs are increasingly central to e-commerce customer interactions. Trained on vast datasets, they excel in generating personalized responses in multi-turn conversations~\cite{zhao2023survey, naveed2023comprehensive, yi2024survey}. 
Abundant studies have been conducted such that interaction optimization by categorizing queries~\cite{nandkumar2024enhancing}, memory architectures enhancement for better conversational continuity~\cite{liu2024llm}, and utilizing customer profiles for tailored responses through services like CHOPS~\cite{shi2024chops}. Open-source frameworks like LangChain are transforming customer service from static FAQs to dynamic, context-aware systems~\cite{pandya2023automating}. LLMs’ ability to adapt via few-shot learning makes them highly effective in fast-evolving e-commerce environments. However, issues like hallucinations and domain adaptation challenges remain, requiring ongoing optimization.

\subsection{LLM-based RL Agent} 

LLM-based agents have become integral to modern intelligent systems, utilizing advanced language generation and understanding to manage complex, multi-turn dialogues and provide personalized responses. In fields like e-commerce, they analyze customer queries to generate accurate answers and optimize interactions. Advanced prompting techniques, such as Chain-of-Thought (CoT)~\cite{wei2022chain} and ReAct~\cite{yao2022react}, enhance LLMs by enabling step-by-step reasoning and task-specific actions. A key advancement is tool use, where models integrate external tools to access real-time information and perform actions beyond their language capabilities. This allows the agents to tackle more complex tasks, improve their decision-making, and broaden their applicability, helping to address challenges like hallucinations and domain adaptation that require ongoing optimization~\cite{karpas2022mrklsystemsmodularneurosymbolic, parisi2022talmtoolaugmentedlanguage, schick2023toolformerlanguagemodelsteach, shen2023hugginggptsolvingaitasks}. While these advancements have significantly enhanced LLM-based agents, further optimization can be achieved by integrating RL techniques, which allow agents to improve performance through continuous learning and feedback.

RL enhances LLM-based agents by allowing them to optimize decision-making and improve task performance over time through feedback~\cite{lee2024llm}. In online RL, agents continuously adjust their strategies based on real-time interactions with the environment, whereas offline RL leverages pre-collected data for optimization, refining agent's behavior without the need for real-time interaction~\cite{agarwal2020optimistic, prudencio2023survey, ghosh2022offline}. This integration provides LLM-based agents with greater adaptability, enabling them to tackle complex tasks more effectively~\cite{qi2024webrl, yun2024pretrained, morad2024language, pang2024knowledgeable}. A promising direction in RL research is framing the problem as sequence modeling, which offers new opportunities for improving both the efficiency and scalability of LLM-based agents. One notable approach is Decision Transformer~\cite{chen2021decision}, which utilizes the simplicity and scalability of Transformer architecture to treat RL as conditional sequence modeling. Unlike traditional RL methods that rely on value functions or policy gradients, Decision Transformer predicts optimal actions by conditioning on the desired return (reward), past states, and actions using a causally masked Transformer. This autoregressive approach allows the model to generate actions that align with the specified return, providing a novel perspective on RL as a sequence prediction problem. 
Based on the definition from~\cite{guo2024large}, MindFlow+ is motivated by world simulation and aims to simulate customer representatives' behavior in the e-commerce customer service domain. The system is structured around an LLM-powered agent with the ability to use external tools and interact with the environment, enabling context-aware, adaptive responses. Agent profiling generated is achieved through both pre-defined and data-derived methods, with a focus on simulating the behavior of customer representatives. Capabilities are continuously refined through feedback from both humans and the environment, with memory-based adjustments that allow the agent to evolve and improve over time.

Another key approach to enhance the performance of LLM-based agents is through Human-in-the-loop RL (HIRL), which integrates human feedback during the learning process. When combined with Human-in-the-loop RL (HIRL), LLM-based agents’ adaptability is further enhanced~\cite{chai2020human, wu2022survey, hejna2023few, christiano2017deep, chen2024efficient, sun2023reinforcement, hu2023aligning}. HIRL incorporates human feedback during the learning process, allowing LLM-based agents to optimize their behavior in real-world applications, where both real-time and offline learning can be utilized to fine-tune (FT) responses~\cite{emmons2021rvs}. This combination is particularly beneficial for domains that require continuous interaction or large-scale datasets, offering significant improvements in handling dynamic environments and complex tasks~\cite{kalusivalingam2020enhancing}.

\begin{figure*}
    \centering
    \includegraphics[width=0.9\linewidth]{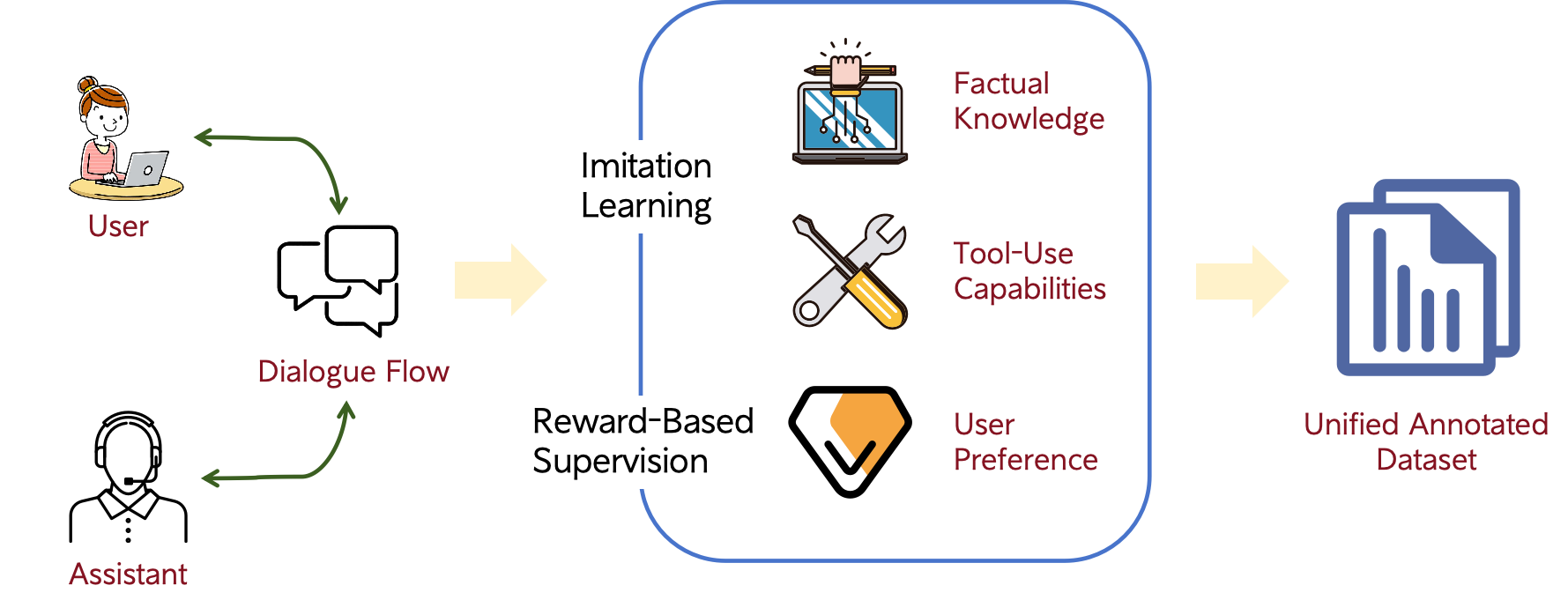}
    \caption{Domain-Specific Training Data Construction Pipeline}
    \label{fig:sft_framework}
\end{figure*}
\section{MindFlow+}

Pre-trained LLMs, while strong in general-purpose tasks, often fall short in delivering context-aware and domain-aligned responses for complex multi-turn dialogue QA. To address this, we introduce MindFlow+, a self-evolving dialogue agent that continuously improves its domain-specific behavior through SFT on a reward-guided corpus. By learning from curated examples that reflect task-specific quality standards and user preferences, MindFlow+ becomes increasingly aligned with practical needs in e-commerce dialogue systems.

Here, we introduce a framework designed to align LLMs with domain-specific behaviors in multi-turn QA settings. Our approach integrates two complementary, data-centric strategies: imitation learning through expert-like demonstrations augmented with knowledge-enhanced demonstrations and ReAct-based agentic demonstrations, empowering the model with factual knowledge and tool-use capabilities; and reward-based supervision, in which user preferences are explicitly annotated using specialized tokens to highlight desirable model responses.

Rather than treating these strategies independently, we unify them at the data construction level to form a reward-aware, tool-augmented training corpus. This hybrid dataset encodes both expert-level task knowledge and preference-aligned behavior signals, enabling the model to learn interaction patterns that reflect both correct domain behavior and downstream feedback. Figure~\ref{fig:sft_framework} illustrates the full pipeline of this domain-specific training data construction.

Through this integrated training process, the resulting model acquires three key capabilities: structured domain knowledge, contextualized tool usage, and fine-grained preference alignment. These enhancements enable even relatively compact models to outperform larger general-purpose counterparts in specialized domains, delivering higher accuracy, faster inference, and reduced computational cost in practical applications.

\subsection{Tool-Augmented Demonstration Construction}

To equip the model with expert-level reasoning and task execution capabilities, we construct imitation data through two complementary augmentation strategies: knowledge-enhanced demonstrations and ReAct-based agentic demonstrations. These strategies enrich raw dialogue data by injecting factual context and demonstrating structured tool-use reasoning.

\subsubsection{Knowledge-Augmented Demonstrations}

Rather than performing retrieval at inference time as in traditional retrieval augmented generation (RAG) systems~\cite{lewis2021rag}, we take a more controlled, training-centric approach. For each multi-turn dialogue instance, relevant background information is heuristically selected from internal knowledge bases and inserted into the system message at the beginning of the conversation. This static context grounding ensures that the model has access to accurate and task-relevant information during response generation.

By conditioning the model on high-quality, domain-specific knowledge without requiring an active retriever component, this approach mitigates hallucination and improves factual consistency, especially in scenarios with stable or repeatable task patterns.

\subsubsection{Agentic Demonstrations via ReAct}

To further enhance the model’s reasoning capabilities and tool-use behaviors, we simulate agent-style decision-making using ReAct prompting~\cite{yao2022react}. Specifically, for each user query, the assistant’s response is structured as a sequence of alternating steps. First, the model formulates a \textit{Thought} by interpreting the user's intent based on the current context and planning how to address the problem. Next, guided by this \textit{Thought}, it takes an \textit{Action} by selecting a specific operation such as issuing a query or invoking an external tool like search, compare, or recommend. Finally, the model receives an \textit{Observation} which is the outcome of the executed action and uses this information for subsequent reasoning and decision-making.


Algorithm~\ref{alg:tool-augmented} illustrate the overall ReAct prompting workflow, where the assistant iteratively alternates between internal reasoning and interactions with external tools in a closed-loop process.

\begin{algorithm*}
\caption{Tool-Augmented Reasoning Loop}
\label{alg:tool-augmented}
\begin{algorithmic}[1]
\Input{Query $query$, History Msgs $hist\_msgs$, Toolset $tools$, Model $model$}
\Output{Final Response $response$}

\State Initialize $observation \gets \emptyset$
\State Initialize $action_{prev} \gets \emptyset$
\While{True}
    \State $thought \gets \text{GenerateThought}(model, query, hist\_msgs, observation, tools)$
    \State $action \gets \text{ExtractAction}(thought)$
    \If{$\text{name}(action) = \text{finish}$}
        \State \textbf{break}
    \EndIf
    \If{$action = action_{prev}$}
        \State \textbf{break}
    \EndIf
    \State $observation \gets \text{Execute}(action)$
    \State $action_{prev} \gets action$
\EndWhile
\State $response \gets \text{GenerateResponse}(model, query, hist\_msgs, observation)$
\State \Return $response$
\end{algorithmic}
\end{algorithm*}

These sequences are embedded directly into the assistant’s turn, allowing the model to observe multiple iterations of decision-making within a single example. Demonstrations are generated via few-shot prompting with open-source LLMs and post-processed for coherence and correctness.

This format equips the model with the ability to emulate real-world workflows: deciding when to use a tool, selecting the appropriate tool for the query, interpreting the returned information, and integrating it into a final response.

Together, these two augmentation strategies form the foundation of our tool-augmented training data. The former exposes the model to grounded factual context, while the latter teaches multi-step decision-making and tool invocation. Importantly, all demonstrations are constructed offline and do not require runtime tool execution, making them suitable for SFT.


 \subsection{Reward-Conditioned Data Modeling}

To enable preference-aware generation without relying on costly online exploration, we adopt a reward-conditioned modeling approach inspired by offline RL. Unlike traditional RL methods requiring environment interaction to estimate reward signals or update policies, offline RL leverages static, pre-collected datasets annotated with feedback signals. This provides a stable and scalable alternative for aligning LLMs to domain-specific or task-specific goals.

We formally model dialogue generation as a Markov Decision Process (MDP) with the tuple $(\mathcal{S}, \mathcal{A}, P, \mathcal{R})$, where states $s \in \mathcal{S}$ correspond to user inputs, actions $a \in \mathcal{A}$ are assistant responses, and $\mathcal{R}(s,a)$ denotes the reward function.

To capture temporal dependencies of multi-turn dialogues, we represent dialogues as interleaved sequences of states, rewards, and actions at each timestep $t$, arranged in a "state–reward–action" order that reflects the causal structure of conversations:

\begin{equation}
\tau = (s_1, r_1, a_1, s_2, r_2, a_2, \ldots, s_t, r_t, a_t)
\end{equation}

This design allows the model to learn how its actions influence subsequent responses and feedback over time.

\begin{figure}[ht]
\centering
\includegraphics[width=1\linewidth]{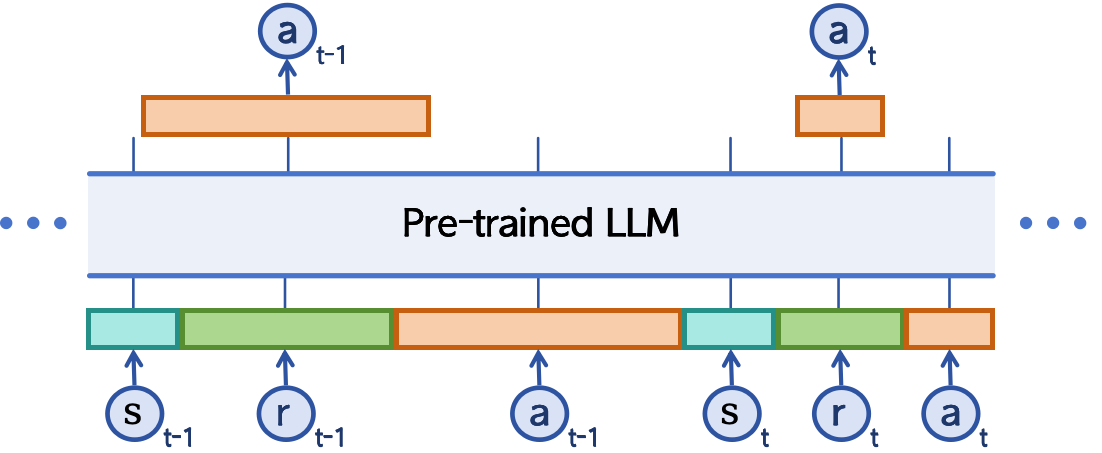}
\caption{Adaptation of Decision Transformer for Reward-Conditioned Dialogue Modeling}
\label{fig:decision_transformer_architecture}
\end{figure}

Inspired by the Decision Transformer (DT) framework~\cite{chen2021decision}, which formulates RL as an autoregressive sequence modeling problem, we adapt this paradigm for dialogue modeling as shown in Figure~\ref{fig:decision_transformer_architecture}. Unlike typical DT applications relying on numeric embeddings, we represent both states and actions as natural language token sequences, enabling seamless integration with LLM pretraining objectives.

To support learning, we construct a reward-annotated dataset where each user query is paired with multiple assistant responses labeled with scalar feedback (e.g., 1 for preferred, 0 for suboptimal). These reward signals are derived from human preferences or automated metrics assessing relevance, factuality, and coherence.

Reward tokens are added to assistant responses as part of the input context, enabling the model to learn preference alignment without modifying the loss function. This offline RL-style setup offers a stable training alternative to traditional policy optimization, supporting context-aware generation in complex multi-turn dialogue tasks.


\subsection{SFT on Tool-Augmented and Reward-Aligned Demonstrations}

SFT provides a stable and scalable framework for adapting pre-trained LLMs to downstream tasks~\cite{zhang2023instruction}. In our approach, SFT is performed over a unified training corpus that integrates both tool-augmented reasoning traces and reward-conditioned responses. This hybrid data structure enables the model to simultaneously learn structured decision-making and preference-aligned generation.

Building upon the previously introduced "state–reward–action" sequence representation, the model is trained to predict the next action given prior states, rewards, and actions by minimizing the standard cross-entropy loss:

\begin{equation}
\mathcal{L}_{\text{SFT}} = - \sum_{t} \log P(a_t \mid s_{\leq t}, r_{\leq t}, a_{<t})
\end{equation}

This formulation treats the reward signal as part of the input context, injected via special tokens such as \texttt{<reward>}, rather than modifying the loss function itself. As a result, the model learns to condition its generation on reward levels while preserving the simplicity and generalization benefits of supervised learning.

The inclusion of both positive and negative responses within the same dialogue context encourages contrastive behavior learning, enabling the model to distinguish high-quality outputs from suboptimal ones. When this training is combined with tool-augmented demonstrations that incorporate structured reasoning and action-observation feedback, the unified SFT process empowers the model to ground its responses in external knowledge through static context or tool observations, perform multi-step, agentic reasoning aligned with user intent, and adapt output quality based on explicit reward feedback.

Overall, this approach bridges long-horizon planning and preference alignment within a single training stage, allowing even moderately sized LLMs to exhibit expert-like behavior in complex, domain-specific QA settings.

\section{Experiments and Evaluation}

\subsection{SFT}

We fine-tune models from the Qwen2.5 series, focusing on those with fewer than 10 billion parameters, to adapt them for complex, multi-turn dialogues in the professional e-commerce domain. These models provide a strong foundation with established language understanding and generation capabilities, enabling efficient SFT tailored to domain-specific tasks.

The SFT process is conducted on 8 NVIDIA A800 GPUs (each with 80GB memory). The maximum input sequence length is set to 4096 tokens. We use a learning rate of $2 \times 10^{-5}$, train for 4 epochs, and adopt an effective batch size of 256.

To better equip the model for practical e-commerce scenarios, we incorporate four domain-specific tools into the dialogue system during training and evaluation:  
\begin{itemize}  
    \item \textbf{Product Information Retrieval Tool}: retrieves detailed specifications and descriptions for queried products.  
    \item \textbf{Product Recommendation Tool}: suggests relevant products based on user preferences and contextual information.  
    \item \textbf{Product Comparison Tool}: provides side-by-side comparisons of multiple products to assist user decision-making.
    \item \textbf{Image Descriptor Tool}: analyzes dialogue-shared images to extract key visual features, enabling enhanced multimodal understanding.  
\end{itemize}  
These tools simulate realistic functionalities commonly used in e-commerce customer service and help the model learn to generate tool-augmented responses, improving its practical utility in multi-turn dialogues.

\subsection{Evaluation}
\subsubsection{Evaluation Metric}
We define the framework for intelligent customer service and highlight the characteristics and application scenarios of each level outlined in the Table. \ref{tab:overview_AI_customer_service}. X1, representing the traditional AI level, features AI that primarily assists human agents, requiring frequent intervention and significant configuration expertise.

\begin{table*}[htbp]
    \centering

    \renewcommand{\arraystretch}{1.5} 
    \begin{tabular}{>{\centering\arraybackslash}m{1cm}>{\centering\arraybackslash}m{3.5cm}>{\centering\arraybackslash}m{5.5cm}>{\centering\arraybackslash}m{4.5cm}}
        \toprule
        \textbf{Rating} & \textbf{Automation Level} & \textbf{Collaboration Models} & \textbf{AI Proficiency Levels} \\
        \midrule
        \textbf{X0} & No Automation & Human-Exclusive & - \\
        \textbf{X1} & Robot Assistance & AI-Assisted, Human-Centric & Entry-Level \\
        \textbf{X2} & Partial Automation & AI-Centric, Human-Assisted & Intermediate; Advanced in Some Scenarios \\
        \textbf{X3} & Conditional Automation & AI Primarily Independent, with Human Supervision for Complex Scenarios & Advanced; Human-Machine Integration at Expert Level \\
        \textbf{X4} & High Automation & AI Fully Independent, with Human Involvement in Training & Expert-Level \\
        \textbf{X5} & Full Automation & AI Autonomous Training and Operation & Beyond Expert-Level \\
        \bottomrule
    \end{tabular}
    
    \caption{Taxonomy of Intelligent Customer Service Standards}
    \label{tab:overview_AI_customer_service}
\end{table*}


To quantify AI's contribution to the customer service process, we introduce the AI Contribution Ratio, calculated as follows:

\begin{equation}
\text{AI Contribution Ratio} = \frac{V_{\text{AI}}}{T_{\text{AI}} + T_{\text{CR}}}
\label{eqa: AI_contribution_ratio}
\end{equation}

where $V_{\text{AI}}$ denotes the number of AI-generated messages judged by customer representatives to be contextually appropriate and relevant within the dialogue. $T_{\text{AI}}$ denotes the total number of AI-generated responses, and $T_{\text{CR}}$ denotes the total number of responses from customer representatives.

The calculation is performed within a defined time window during which messages are monitored. This metric provides a clear indication of AI's participation and automation level across various collaboration models, offering insights into how AI’s role evolves from X1 to higher levels. Unlike traditional evaluation metrics such as accuracy or F1 score, the AI Contribution Ratio evaluates the model’s specific contribution to user decision-making and task completion in real-world dialogues. It considers not only the accuracy of the generated content but also its practical value in the e-commerce context. By assessing performance across multi-turn dialogues, this metric measures adaptability and collaborative effectiveness, providing deeper insights into the model's performance in e-commerce question-answering tasks.

To efficiently evaluate model performance, we automate the scoring process using a preset prompt that incorporates customer inquiries, customer representatives' responses, and MindFlow+ outputs (which are the AI-generated messages). We invoke an LLM API to score these outputs (denoted as 0 for Invalid AI Messages and 1 for Valid AI Messages). To ensure alignment with manual scoring, we label a small test set and calculate the Spearman's correlation between manual and API scores. This correlation is used to iteratively refine the prompt, ensuring statistical significance between API and manual scores~\cite{dubois2024length}.


\subsubsection{Experimental Setting} 
We conduct our evaluation on 150 real-world user queries collected from an e-commerce customer service platform. The conversations originate from a store selling products in a representative category (e.g., consumer electronics), covering diverse scenarios such as parameter inquiries, setup assistance, greetings, product recommendation, product comparison, multimodal understanding, and multi-turn dialogues that often require deeper user intent interpretation. The main challenge lies in the complexity and diversity of the queries, many of which require reasoning over background knowledge, identifying user needs over multiple turns, and generating helpful, context-aware responses.

\subsubsection{Experimental Results}

Table~\ref{tab:experimental_results} summarize the evaluation results. Baseline model is trained on dialogues augmented with relevant background information, such as store descriptions and product names. This information encodes key product attributes and enhances the informativeness of each turn. As a result, the model achieves reasonable performance even without task-specific alignment.

The Tool-Augmented model is trained to follow chain-of-thoughts and invoke tools appropriately. However, the original dataset's answer annotations often mimic realistic service responses containing delays or vague phrases (e.g., "please wait", "I can't help with that"), leading to poor response quality despite learned tool-calling behaviors.

\begin{table*}[]
    \centering
    \resizebox{0.8\textwidth}{!}{
    \renewcommand{\arraystretch}{1.5}
    \begin{tabular}{ccccc}
    \specialrule{1.2pt}{0pt}{0pt}
      \textbf{Pretrained Models}     &  \textbf{Baseline} & \textbf{Tool-Augmented} & \textbf{Reward-Guided} & \textbf{MindFlow+} \\
        \hline
      Qwen2.5-0.5B-Instruct   &  35.33\% & 4.00\%  & 66.67\% & \textbf{72.67\%} \\
      Qwen2.5-1.5B-Instruct   &  44.00\% & 4.67\%  & 61.33\% & \textbf{85.33\%}  \\
      Qwen2.5-3B-Instruct     &  56.00\% & 6.67\%  & 57.33\% & \textbf{87.33\%} \\
      Qwen2.5-7B-Instruct     &  58.00\% & 14.00\% & 72.67\% & \textbf{94.00\%} \\
      \specialrule{1.2pt}{0pt}{0pt}
    \end{tabular}}
    \caption{Evaluation Results on AI Contribution Ratio}
    \label{tab:experimental_results}
\end{table*}

The Reward-Guided model is trained with both positive and negative response samples. Although it does not invoke tools, it has access to the same background information as the baseline, and learns to generate responses aligned with expert-level behavior, achieving notable improvements.

MindFlow+ model combines both reward-guided and tool-augmented signals. It learns to both make appropriate tool calls and generate effective, preference-aligned responses. This dual ability leads to superior performance.

\subsection{Ablation Study and Discussion}

In this section, we conduct a series of ablation experiments to examine the effects of different modeling choices on the AI contribution ratio. Specifically, we analyze the impact of the reward token position and compare multiple pretraining strategies.

\subsubsection{Effect of Reward Token Position}

\begin{table}[]
    \centering
    
    \renewcommand{\arraystretch}{1.5}
    \begin{tabular}{ccc}
    \specialrule{1.2pt}{0pt}{0pt}
      \textbf{Pretrained Models} & $\boldsymbol{\{s,r,a\}}$ order & $\boldsymbol{\{r,s,a\}}$ order\\
        \hline
      Qwen2.5-0.5B-Instruct   & \textbf{66.67\%} & 33.33\%\\
      Qwen2.5-1.5B-Instruct   & \textbf{61.33\%} & 15.33\%\\
      Qwen2.5-3B-Instruct     & \textbf{57.33\%} & 52.00\%\\
      Qwen2.5-7B-Instruct     & \textbf{72.67\%}& 48.00\%\\
      \specialrule{1.2pt}{0pt}{0pt}
    \end{tabular}
    \caption{Comparison of Reward Token Position}
\label{tab:comparison_reward_token_position}
\end{table}

We adopt a standardized reward token format denoted as \verb+<reward>score</reward>+ for all reward-aligned demonstrations. To investigate the impact of reward token placement, we compare two sequence configurations: $\{s,r,a\}$ (state → reward → action) and $\{r,s,a\}$ (reward → state → action). The results are summarized in Table~\ref{tab:comparison_reward_token_position}.


Unlike Decision Transformers that operate over numeric trajectories, our LLM-based model is sensitive to sequence order. The above results suggest that the $\{s,r,a\}$ configuration yields better alignment, likely due to the proximity of state and action tokens reinforcing reward semantics via the attention mechanism.


\subsubsection{Comparison of Pretraining Strategies}

We compare three pretraining setups: (1) Baseline, which includes background information such as product descriptions and store attributes; (2) Baseline w. Tool, which introduces explicit tool usage prompts in the input to test prompt-based tool reasoning capabilities; and (3) Background-Free, which excludes background knowledge but retains user and assistant roles.

As shown in 
Table~\ref{tab:pretraining_comparison}, the Baseline model performs reasonably well, supported by accessible background features. The Background-Free setting highlights the performance drop when factual context is unavailable, revealing the importance of knowledge-augmented demonstrations. Meanwhile, the Tool-Prompt baseline shows limited improvement, indicating that pretraining alone is insufficient for tool learning without supervision or reward guidance.

\begin{table*}[]
    \centering
    \resizebox{0.6\textwidth}{!}{
    \renewcommand{\arraystretch}{1.5}
    \begin{tabular}{cccc}
    \specialrule{1.2pt}{0pt}{0pt}
      \textbf{Pretrained Models}     &  \textbf{Baseline} & \textbf{Baseline w. Tool} & \textbf{Background-Free} \\
        \hline
      Qwen2.5-0.5B-Instruct   &  \textbf{35.33\%} & 26.00\% & 28.67\%\\
      Qwen2.5-1.5B-Instruct   &  \textbf{44.00\%} & 38.00\% & 25.33\%\\
      Qwen2.5-3B-Instruct     &  \textbf{56.00\%} & 36.67\% & 28.00\%\\
      Qwen2.5-7B-Instruct     &  \textbf{58.00\%} & 28.00\% & 37.77\%\\
      \specialrule{1.2pt}{0pt}{0pt}
    \end{tabular}}
    \caption{Effect of Pretraining Strategies}
    \label{tab:pretraining_comparison}
\end{table*}

\subsection{Cross-Domain Generalization}

To evaluate the generalizability of our proposed model trained on the unified-annotated dataset, we apply it directly to a new domain, a store that sells household essentials, without any additional training, prompt engineering, or domain-specific adaptation.

We randomly select 30 real-world conversations from this new store for evaluation. As shown in Table~\ref{tab:household_generalization}, the AI contribution ratio remains consistently high across different model scales, demonstrating the robustness and reusability of our method in unseen scenarios.

\begin{table}[]
    \centering
    
    \renewcommand{\arraystretch}{1.5}
    \begin{tabular}{cc}
    \specialrule{1.2pt}{0pt}{0pt}
      \textbf{Pretrained Models}     &  \textbf{AI Contribution Ratio} \\
        \hline
      Qwen2.5-0.5B-Instruct   &  73.33\% \\
      Qwen2.5-1.5B-Instruct   &  80.00\% \\
      Qwen2.5-3B-Instruct     &  86.67\% \\
      Qwen2.5-7B-Instruct     &  90.00\% \\
      \specialrule{1.2pt}{0pt}{0pt}
    \end{tabular}
    \caption{AI Contribution Ratio On A New Domain Without Any Further Tuning}
    \label{tab:household_generalization}
\end{table}

Overall, these results demonstrate that our method is both robust and reusable, with the model exhibiting strong generalization capabilities in unseen domains. This indicates that its behavior can effectively transfer to structurally similar customer service environments with minimal adaptation effort.

\section{Conclusion}

We present MindFlow+, a self-evolving agent framework tailored for e-commerce customer service. MindFlow+ acquires domain-specific capabilities through SFT on interaction traces enriched with tool calls, external knowledge, and reward-guided, preference-aligned supervision at the token level. These enriched signals support alignment with task-specific goals and enable the agent to dynamically interact with external tools and incorporate domain-specific knowledge in a controllable and interpretable manner.

To assess real-world performance, we introduce the AI Contribution Ratio metric to quantify agent autonomy in tool-assisted workflows. Extensive experiments on real-world customer service data from a large-scale e-commerce platform validate the effectiveness of MindFlow+.

While MindFlow+ demonstrates strong alignment and robustness, limitations remain. The current pipeline is tailored to specific toolsets and domains, and future work will explore generalization across verticals, including low-resource or zero-shot settings. We also plan to extend MindFlow+ to high-stakes applications such as healthcare and finance, where safety, reliability, and alignment are even more critical.

\appendices
\section{\break Data Formatting}
\begin{enumerate}
    \item \textbf{Original Paired QA}\\Standard user–assistant QA pairs\\
    \begin{framed}
    \begin{lstlisting}
{"role": "user", "content": "query"}
{"role": "assistant", "content": "original answer"}
\end{lstlisting}
    \end{framed}
    
    \item \textbf{Knowledge-Augmented Demonstrations (Baseline)}\\Dialogues augmented with relevant background context
    \begin{framed}
    \begin{lstlisting}
{"role": "system", "content": "conversation background"}
{"role": "user", "content": "query"}
{"role": "assistant", "content": "original answer"}
\end{lstlisting}
    \end{framed}

    \item \textbf{Agentic Demonstrations via ReAct}\\Baseline with reasoning and action steps in assistant responses
    
    \begin{framed}
\begin{lstlisting}
{"role": "system", "content": "conversation background"}
{"role": "user", "content": "query"}
{"role": "assistant", "content": "<thought>...</thought><action>...</action><observation>...</observation><answer>original answer</answer>"}
    \end{lstlisting}
    \end{framed}

    \item \textbf{Reward-Conditioned Data Modeling}\\Baseline with multiple responses per query annotated with scalar rewards
    
    \begin{framed}
    \begin{lstlisting}
    {"role": "system", "content": "conversation background"}
{"role": "user", "content": "query"}
{"role": "assistant", "content": "<reward>1</reward>preferred answer"}
{"role": "assistant", "content": "<reward>0</reward>suboptimal answer"}
    \end{lstlisting}
    \end{framed}

    \item \textbf{Integrated Reasoning-Reward Demonstrations}\\Baseline combining reasoning traces and reward annotations
    
    \begin{framed}
    \begin{lstlisting}
{"role": "system", "content": "conversation background"}
{"role": "user", "content": "query"}
{"role": "assistant", "content": "<thought>...</thought><action>...</action><observation>...</observation><reward>1</reward>preferred answer"}
{"role": "assistant", "content": "<thought>...</thought><action>...</action><observation>...</observation><reward>0</reward>suboptimal answer"}
    \end{lstlisting}
    \end{framed}
\end{enumerate}

\section{\break Chat Template}

To ensure compatibility with the pretrained tokenizer and conversation prompt design, we adopt the ChatML format. Each message is wrapped with \verb+<|im_start|>+ and \verb+<|im_end|>+ tags, which clearly delineate conversational turns. This format preserves role-based structure and supports variable-length reasoning blocks.

The following example illustrates how a tool-augmented, reward-scored assistant response is formatted within the ChatML framework. It presents a hypothetical academic research assistant scenario where the model helps users find relevant scientific literature, explains complex concepts, and provides thoughtful, context-aware answers. The dialogue showcases the integration of external knowledge from the system prompt and the use of the ReAct paradigm for reasoning and tool-based actions during multi-turn interactions. Additionally, the model generates responses aligned with predefined reward scores, ensuring high-quality and usable outputs.

\begin{framed}
\verb+<|im_start|>system+

You are an expert academic research assistant, skilled in helping users find relevant scientific papers, explaining complex concepts, and providing thoughtful answers. Below is the background information for this session: ...

\verb+<|im_end|>+

\verb+<|im_start|>user+

Can you help me find papers about graph neural networks (GNNs) applied in chemistry?

\verb+<|im_end|>+

\verb+<|im_start|>assistant+

\verb+<thought>+ The user is requesting scientific literature on GNNs in chemistry. I should search relevant databases. \verb+</thought>+

\verb|<action><tool_call>|{"name":"Search", "arguments": {"query": "graph neural networks chemistry"}}\verb|</tool_call></action>|

\verb|<observation>| Found several papers, including "Molecular Property Prediction with Graph Neural Networks". \verb|</observation>|

\verb|<reward>1</reward>| Sure! One relevant paper is "Molecular Property Prediction with Graph Neural Networks". It demonstrates how GNNs can predict molecular properties effectively.

\verb+<|im_end|>+

\end{framed}

The integration of ChatML and structured special tokens equips the model with the ability to perform multi-step reasoning, invoke tools, and align responses with reward signals, all within a unified fine-tuning framework.

\bibliographystyle{IEEEtran}   
\bibliography{MindFlow}     

\begin{IEEEbiography}[{\includegraphics[width=1in,height=1.25in,clip,keepaspectratio]{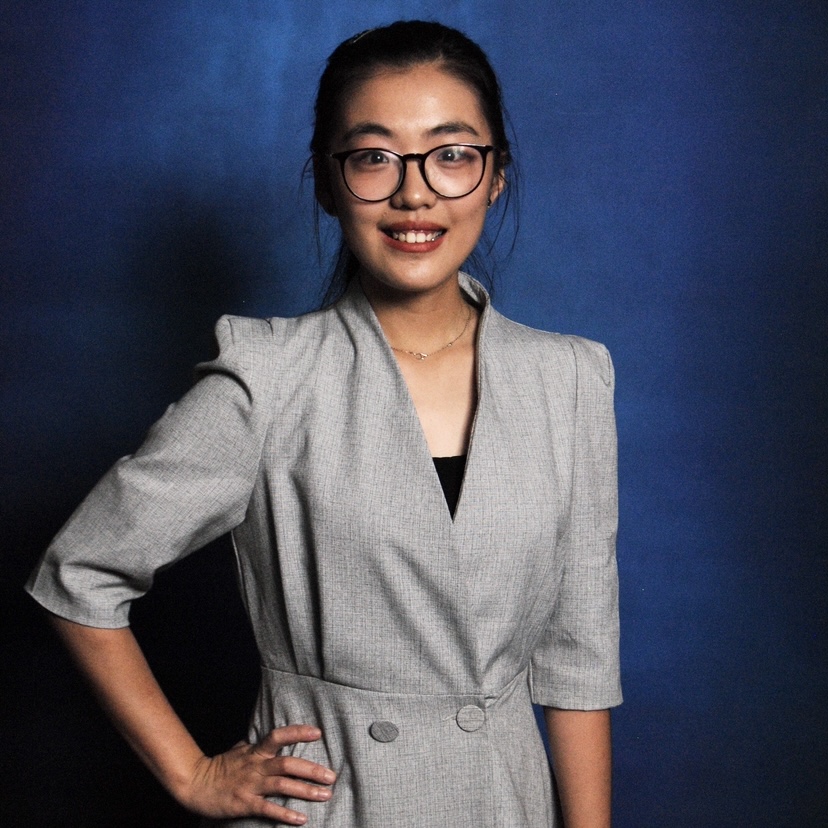}}]{Ming Gong}
received the M.S. and Ph.D. degrees in Electrical Engineering from the University of Dayton. She was previously affiliated with the Vision Lab at the University of Dayton, where she conducted research in computer vision, including object recognition, line segment detection, and semantic segmentation, and published multiple papers in these areas. She is currently an intern at the Xiaoduo AI Lab, focusing on natural language processing, large language models (LLMs), agent frameworks, and multimodal LLMs, with recent publications in these domains.
\end{IEEEbiography}

\begin{IEEEbiography}[{\includegraphics[width=1in,height=1.25in,clip,keepaspectratio]{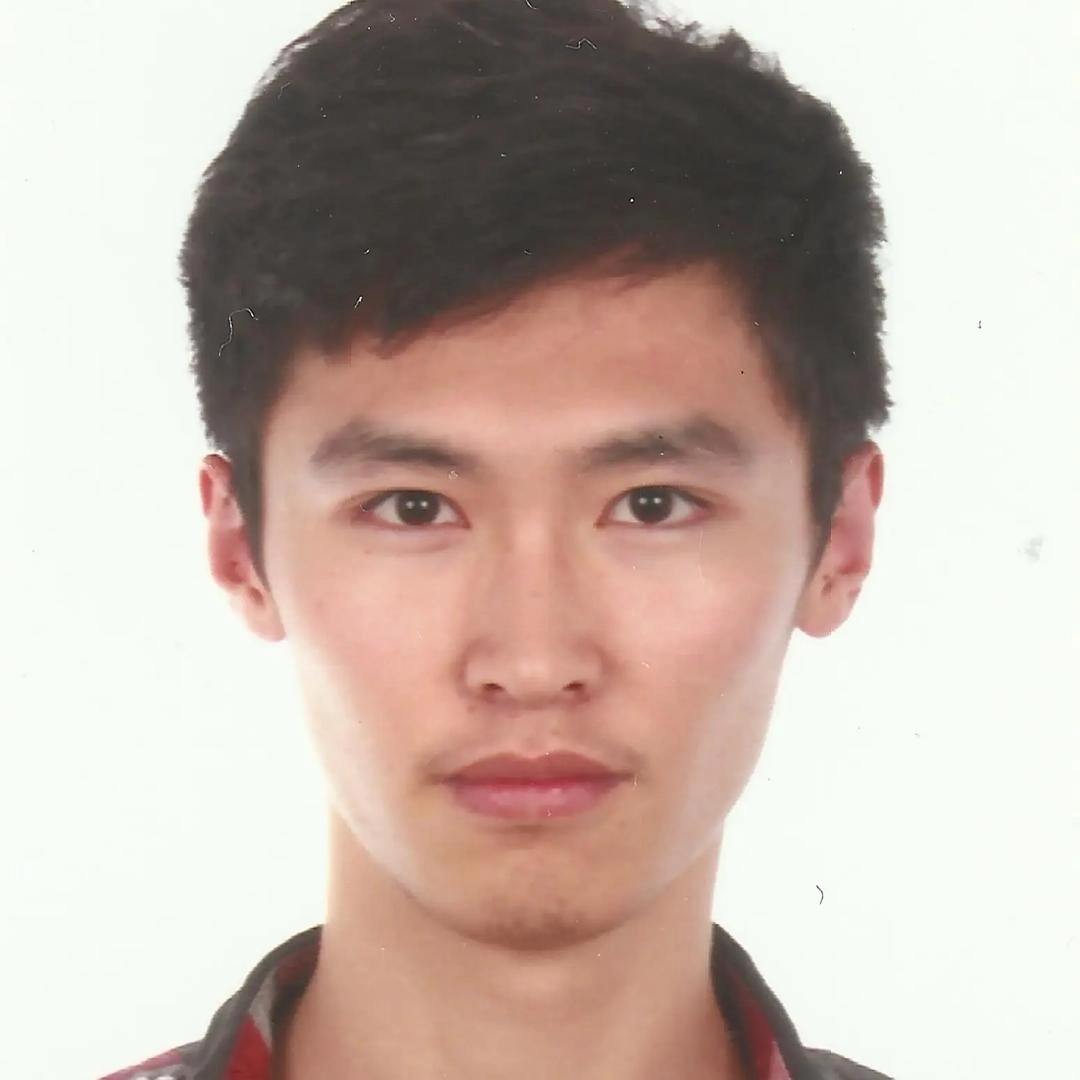}}]{Xucheng Huang} 
served as a Research Assistant at the Xiaoduo \& Aiit Union AI Laboratory from 2023 to 2025, he. Since 2019, he has been working as an student assistant in the Fifth Information Department at Friedrich-Alexander University Erlangen-Nürnberg (FAU). He has participated in two notable projects: DIBCO handwritten document image denoising, and medical acoustic and linguistic analysis for Alzheimer's disease. His research interests include image instance segmentation, image denoising, text classification, NER, spell correction, LLM pre-training, and multimodal LLM information comprehension.
\end{IEEEbiography}

\begin{IEEEbiography}[{\includegraphics[width=1in,height=1.25in,clip,keepaspectratio]{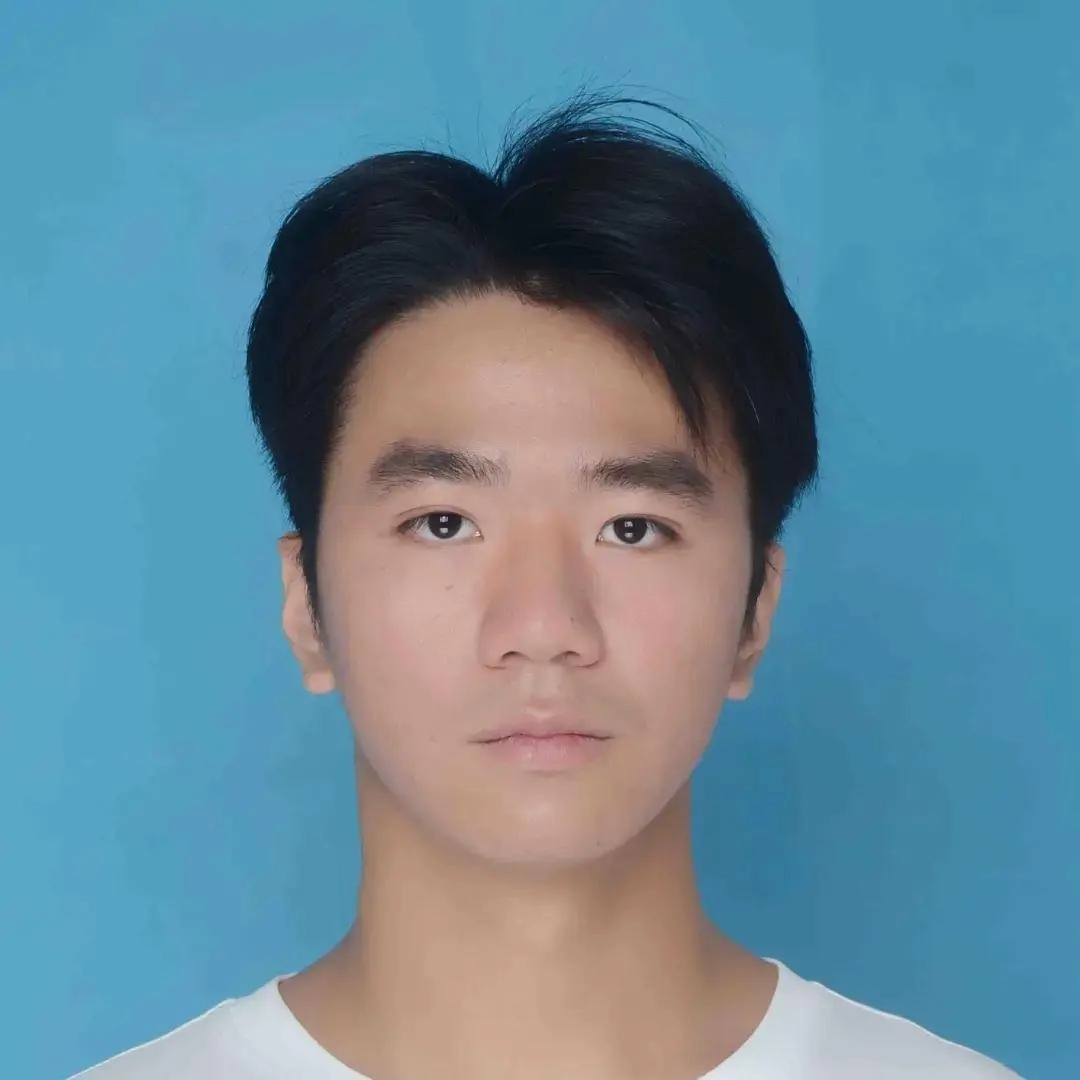}}]{Ziheng Xu} 
received the B.S. degree in Geographic Information Science from East China Normal University in 2022 and the M.S. degree in Geomatics Engineering from the same university in 2025. During 2023-2024, he served as an NLP Algorithm Intern at the AI Department of Banma. From 2024 to 2025, he was an Algorithm Intern at the AI Department of Xiaoduotech.
He has published three papers, with primary research interests including:
Human-Computer Interaction intelligent cockpits,E-commerce AI customer service.
\end{IEEEbiography}

\begin{IEEEbiography}[{\includegraphics[width=1in,height=1.25in,clip,keepaspectratio]{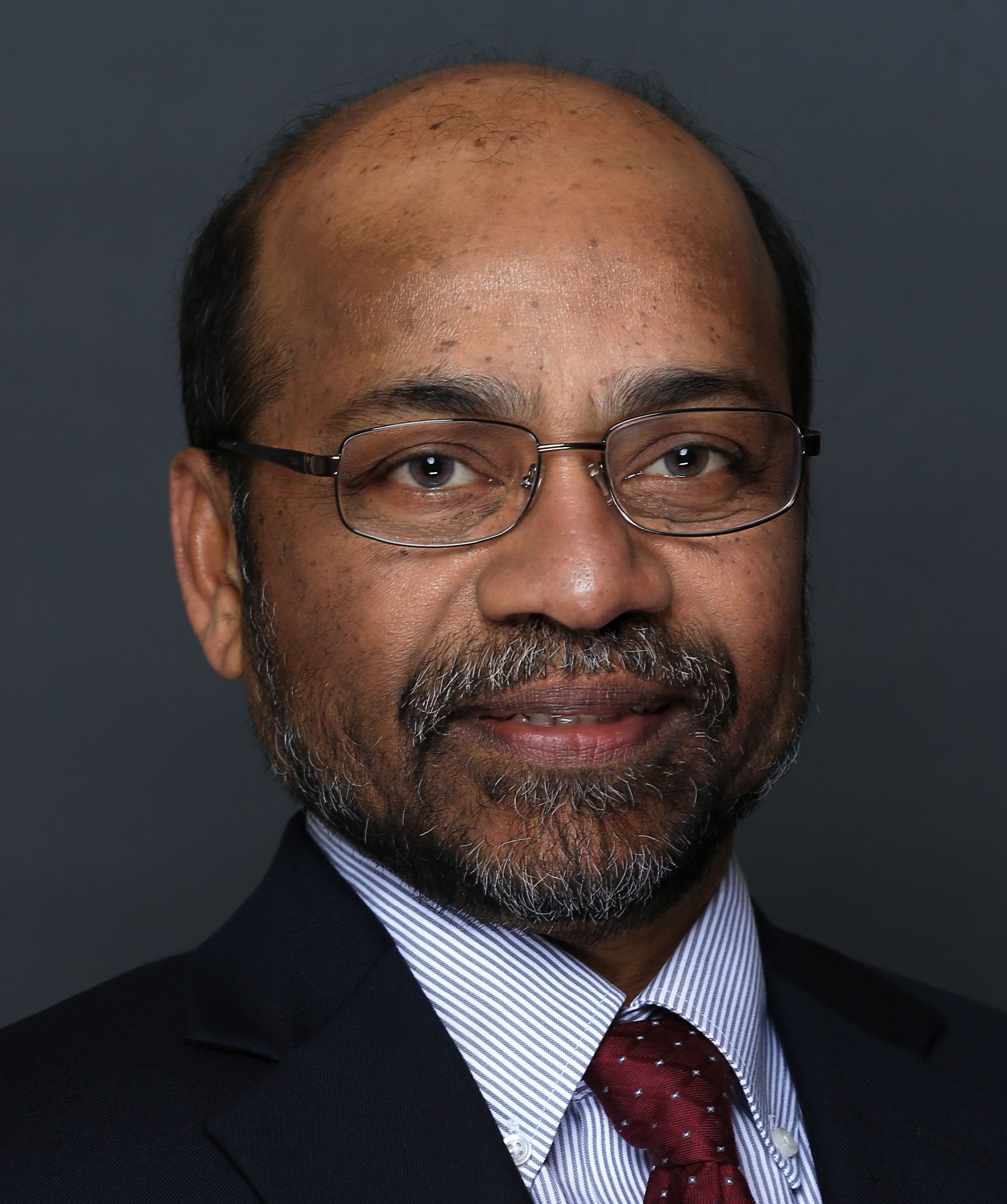}}]{Dr. Vijayan K. Asari} 
is a Professor in Electrical and Computer Engineering and Ohio Research Scholars Endowed Chair in Wide Area Surveillance at University of Dayton. He is the director of the Center of Excellence for Computational Intelligence and Machine Vision at UD.  Dr. Asari received his PhD degree in Electrical Engineering from Indian Institute of Technology, Madras. Prior to joining UD in February 2010, Dr. Asari worked as Professor in ECE at Old Dominion University, Norfolk, Virginia. Dr. Asari holds five patents and has published more than 800 research articles, including 142 peer-reviewed journal papers in the areas of image processing, pattern recognition, machine learning and artificial neural networks. He has so far mentored 33 PhD dissertations and 50 MS theses in electrical and computer engineering. Dr. Asari received several awards for teaching, research, advising, and technical leadership. He is an elected Fellow of SPIE and a Senior Member of IEEE.
\end{IEEEbiography}

\EOD

\end{document}